\documentclass[conference]{IEEEtran}
\IEEEoverridecommandlockouts
\usepackage{cite}
\usepackage{amsmath,amssymb,amsfonts}
\usepackage{algorithmic}
\usepackage{graphicx}
\usepackage{textcomp}
\usepackage{xcolor}
\def\BibTeX{{\rm B\kern-.05em{\sc i\kern-.025em b}\kern-.08em
    T\kern-.1667em\lower.7ex\hbox{E}\kern-.125emX}}

\usepackage{balance} % balance the references
\usepackage{mathtools}
\usepackage[colorlinks=true, urlcolor=blue, linkcolor=red]{hyperref}

\usepackage{diagbox} % for table drawing

\begin{document}

\makeatletter
\newcommand{\linebreakand}{%
  \end{@IEEEauthorhalign}
  \par
  % \hfill\quad\mbox{}\begin{@IEEEauthorhalign}
    % \hfill\mbox{}\par
    \mbox{}\hfill\begin{@IEEEauthorhalign}
}
\makeatother

\setlength{\textfloatsep}{0.7\baselineskip plus 0.2\baselineskip minus 0.5\baselineskip}
\setlength{\abovecaptionskip}{2pt}
\setlength{\belowcaptionskip}{5pt}

\title{A Vision-based Autonomous Perching \\ Approach for Nano Aerial Vehicles \\
\thanks{*Corresponding author.}
}

\author{
\IEEEauthorblockN{Truong-Dong Do}
\IEEEauthorblockA{\textit{Dept. Aerospace Systems Engineering} \\
\textit{Dept. of Convergence Engineering} \\
\textit{for Intelligent Drone}\\
\textit{Sejong University}\\
Seoul, South Korea \\
dongdo@sju.ac.kr}
\and
% \IEEEauthorblockN{Nguyen Xuan-Mung}
% \IEEEauthorblockA{\textit{Faculty of Mechanical and} \\
% \textit{Aerospace Engineering} \\
% \textit{Sejong University}\\
% Seoul, South Korea \\
% xuanmung@sejong.ac.kr}
% \and
% \IEEEauthorblockN{Hansol Jeong}
% \IEEEauthorblockA{\textit{Dept. Aerospace Systems Engineering} \\
% \textit{Dept. of Convergence Engineering} \\
% \textit{for Intelligent Drone}\\
% \textit{Sejong University}\\
% Seoul, South Korea \\
% minisol0175@gmail.com}
% \linebreakand % <------------- \and with a line-break to auto align
% % \and
% \IEEEauthorblockN{\hspace{0.7cm}Yong-Seok Lee}
% \IEEEauthorblockA{\hspace{0.7cm}\textit{Dept. Aerospace Systems Engineering} \\
% \textit{\hspace{0.7cm}Dept. of Convergence Engineering} \\
% \textit{\hspace{1cm}for Intelligent Drone}\\
% \textit{\hspace{1cm}Sejong University}\\
% \hspace{1cm}Seoul, South Korea \\
% \hspace{1cm}drag97@naver.com}
% \and
% % \linebreakand
% \IEEEauthorblockN{Chang-Woo Sung}
% \IEEEauthorblockA{\textit{Dept. Aerospace Systems Engineering} \\
% \textit{Dept. of Convergence Engineering} \\
% \textit{for Intelligent Drone}\\
% \textit{Sejong University}\\
% Seoul, South Korea \\
% cw\_0728@naver.com}
% \and
\IEEEauthorblockN{Sung Kyung Hong*}
\IEEEauthorblockA{\textit{Faculty of Mechanical and} \\
\textit{Aerospace Engineering} \\
\textit{Dept. of Convergence Engineering} \\
\textit{for Intelligent Drone}\\
\textit{Sejong University}\\
Seoul, South Korea \\
skhong@sejong.ac.kr}
}

\IEEEaftertitletext{\vspace{-2\baselineskip}}

% \IEEEoverridecommandlockouts
% \IEEEpubid{\makebox[\columnwidth]{979-8-3503-2294-1/23/\$31.00~\copyright2023 IEEE \hfill}
% \hspace{\columnsep}\makebox[\columnwidth]{ }}

\maketitle

% \IEEEpubidadjcol

\begin{abstract}
Over the past decades, quadcopters have been investigated, due to their mobility and flexibility to operate in a wide range of environments. They have been used in various areas, including surveillance and monitoring. During a mission, drones do not have to remain active once they have reached a target location. To conserve energy and maintain a static position, it is possible to perch and stop the motors in such situations. The problem of achieving a reliable and highly accurate perching method remains a challenge and promising. In this paper, a vision-based autonomous perching approach for nano quadcopters onto a predefined perching target on horizontal surfaces is proposed. First, a perching target with a small marker inside a larger one is designed to improve detection capability at a variety of ranges. Second, a monocular camera is used to calculate the relative poses of the flying vehicle from the markers detected. Then, a Kalman filter is applied to determine the pose more reliably, especially when measurement data is missing. Next, we introduce an algorithm for merging the pose data from multiple markers. Finally, the poses are sent to the perching planner to conduct the real flight test to align the drone with the target's center and steer it there. Based on the experimental results, the approach proved to be effective and feasible. The drone can successfully perch on the center of markers within two centimeters of precision.

\end{abstract}


\section{Introduction}
Unmanned Aerial Vehicles (UAVs), including drones and multicopters, are becoming popular research subjects due to their maneuverability and autonomy \cite{silvagni2017multipurpose,tran2018vision,xuan2022quadcopter,lee2023adaptive}. With the growing popularity of UAVs, it is necessary to increase awareness of the environment while improving flight performance. The small size and light weight of nano UAVs make them ideal platforms \cite{giernacki2017crazyflie,garcia2017modeling,nguyen2022quadrotor,niculescu2021improving}. However, they have a very limited flight time. Fortunately, most missions do not require hovering for the entire duration. Therefore, it is necessary to develop autonomous perching solutions to conserve energy.

Perching refers to supporting the aerial robot's weight from within using grasping, attachment, or embedding techniques \cite{meng2022aerial}. Furthermore, this capability could be useful for tasks that require robots to maintain precise, static positions, function as radio relays during disasters, or suspend operations in unfavorable weather conditions. Considering the surface to be reached by the vehicle, solving the perching control problem requires more effort than addressing the landing task \cite{mao2023robust}. Besides, the perching control performance can be significantly degraded due to several reasons, such as sensor shortcomings and external disturbances.

Among various sensors that can be used in solving this problem, visual cameras are the class used widely the most \cite{10003944,do2018real,tran2021enhancement}. These camera modules provide simple, affordable, and reliable solutions that can greatly improve UAV navigation systems. In the field of visual servoing, there is foundational literature covering control using monocular vision, which discusses the differences between Position Based Visual Servoing (PBVS) and Image-Based Visual Servoing (IBVS) \cite{hutchinson1996tutorial, chaumette2006visual}. The visual servoing approaches \cite{thomas2015visual,zhang2018optimal} have shown autonomous perching results without the use of motion capture but are highly dependent on objects’ shapes and require the object to initially be in the field of view.

% Add figure
\begin{figure*}[t!]
\centering
\includegraphics[width=0.8\textwidth]{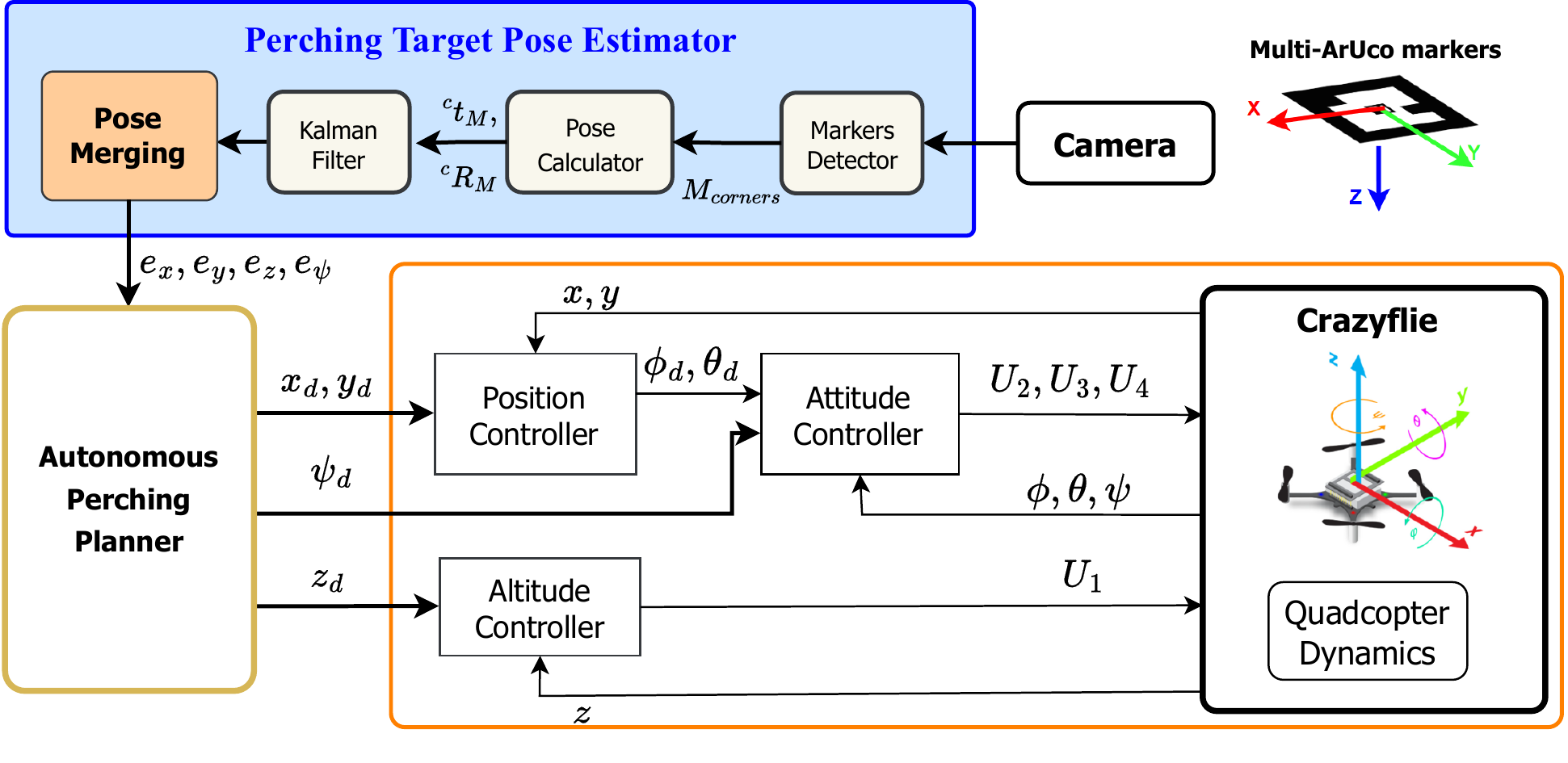}
\vspace*{-3mm}
\caption{Diagram of the control structure of the proposed autonomous perching algorithm. The components contained within the orange line define operations that perform in the Crazyflie firmware.}
\vspace{-5mm}
\label{fig:fig_1}
\end{figure*}

The use of the visual camera is usually accompanied by one or more makers. Large markers are good for the distant detection of the camera. However, when the sensor and the markers are close enough, the markers can be out of the sensor’s field of view causing the target loss phenomenon and the perching failure. In contrast, a small marker offers the advantage of being detectable when the drone approaches, but it is difficult to identify from a distance\cite{10003944,do2023vision}.

In this research, a vision-based autonomous perching method for nano drones onto a specified perching target on horizontal surfaces is proposed to deal with the above-mentioned problems. First, the perching target is designed to enhance detection capability at both a wide and close distance. Second, the pose of ArUco markers \cite{garrido2014automatic} was estimated. Third, a Kalman filter is applied to provide a more accurate and consistent pose, especially when measurement data is missing due to transmission losses, camera vibrations, or missing detection data. Next, we introduce a merging algorithm as well as the perching planner for high-precision perching performance. Finally, experiments with real flight tests to validate the strategy were conducted. The preliminary results indicate the effectiveness and feasibility of our proposal for autonomous nano-UAVs perching research.

The remainder of this article is organized as follows. Section \ref{Methodology} provides an overview of the methodologies used to accomplish the task. The experiments, including the system's configuration and the obtained results, are presented in Section \ref{Experiments}. Finally, in Section \ref{Conclusion}, the paper concludes by outlining future research and development opportunities.

% Add figure
\begin{figure*}[ht!]
\centering
\includegraphics[width=0.6\textwidth]{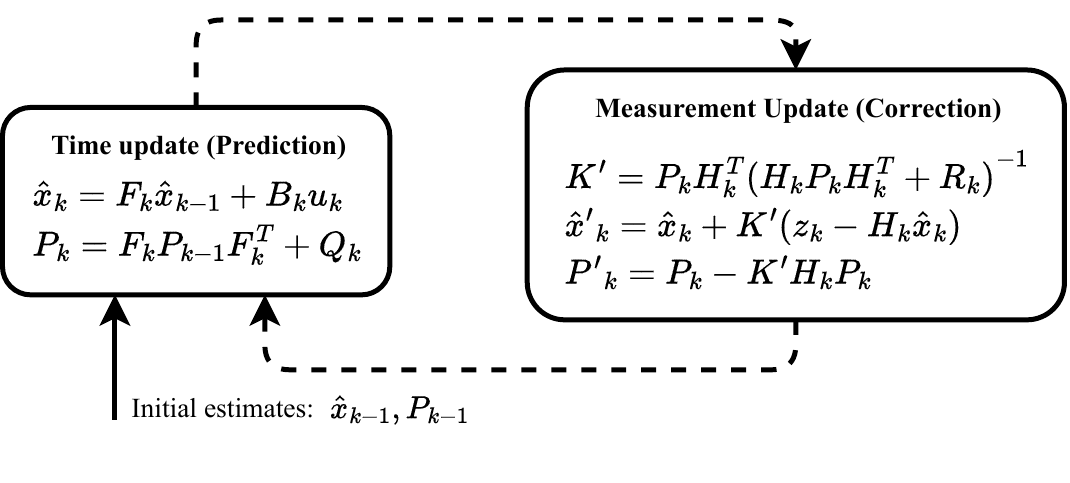}
\vspace*{-6mm}
\caption{Operation of Kalman Filter.}
\vspace{-5mm}
\label{fig:kalman}
\end{figure*}

\section{Methodology} \label{Methodology}
The diagram of the control structure of the proposed autonomous perching algorithm is demonstrated in Fig. \ref{fig:fig_1}. In order to use the ArUco library for detecting markers and estimating poses, the camera must first be calibrated to be able to determine the camera's pose in the scene accurately and retrieve information about their size in real-world units. 

Through the ArUco library from OpenCV \cite{bradski2000opencv}, each obtained image frame $(J)$ is processed in the computer, and once the markers ($M_{corners}$) has been detected by the algorithm, the pose of the quadcopter relative to each marker (${}^{c}{R}_{M}$, ${}^{c}{{t}_{M}}$) is calculated by Perspective-n-Point ($PnP$) solver algorithm \cite{lepetit2009ep} and is fed to the Kalman filter \cite{kalman1960new}. The pose estimator outputs the relative pose error between the target and drone ($e_{x}$, $e_{y}$, $e_{z}$, $e_{\psi}$) in the world frame centimeters and degrees, respectively, as the input of perching planner.

After the perching planner produces its output, Crazyflie \cite{bitcraze21} aligns itself toward the target as desired. The control signals are comprised of the relative $yaw$-angle $\psi_{d}$ and translation in $x$, $y$, and $z$-axis ($x_{d}$, $y_{d}$, $z_{d}$). The $x$-axis defines the forward and backward direction, the $y$-axis defines the left and right, $z$-axis defines the vertical upward to the target. The onboard controller of Crazyflie firmware includes controllers (position, altitude, and attitude) in cascade which get the desired inputs from the perching planner to calculate the final control outputs ($U_{1}$, $U_{2}$, $U_{3}$, $U_{4}$).

\subsection{Proposed Perching Target Pose Estimator}

The perching target, including a small marker located inside a large one as in Fig. \ref{fig:fig_1}. A larger maker's size will improve detection accuracy but will also increase computational complexity and processing time. Thus, we chose simple markers to detect in accordance with the requirement for quickness. The larger square marker has a size of \(150mm\) and belongs to the $DICT\_4x4\_100$ dictionary with the $ID = 997$. Meanwhile, to prevent the detect confusion, the smaller marker is chosen from $DICT\_ARUCO\_ORIGINAL$ with $ID$ = 5. It is placed at the center of the larger marker with $25mm$ in the dimension. Both have the same origin and coordinate; adding the smaller marker may harm the detection of the larger one. However, this problem did not arise in our tests.

\subsubsection{\textbf{Estimate the missing data with Kalman filter}}
Considering the limitations of the camera, transmission losses, or loss of detection, it requires a method for estimating the relative pose of the drone given the incomplete or noisy data from the images. The Kalman filter cleans up the measured data and projects the measurement onto a state estimate. It addresses the problem of predicting the state of a dynamical system at a discrete-time step $k$, given measurements from the current state at the time step $k-1$ and its uncertainty matrix.

The data that we are looking to estimate and filter includes the relative $yaw$-angle  and the translation between the body-fixed frame of the quadrotor and the reference frame of the detected marker with the state vector: ${}^{c}{{P}_{M}}=\left\{ \psi,\dot{\psi },{{t}_{x,}}{{{\dot{t}}}_{x,}}{{t}_{y,}}{{{\dot{t}}}_{y,}}{{t}_{z,}}{{{\dot{t}}}_{z,}} \right\}$, where $c$, $M$ is the camera and markers reference frame respectively. The operation of the Kalman filter is illustrated in Fig. \ref{fig:kalman}. Having a state vector containing eight elements implies that we need a state transition vector initially defined as:
\begin{equation}
{{F}_{k}}=\left[ \begin{matrix}
   1 & \Delta t & 0 & 0 & 0 & 0 & 0 & 0  \\
   0 & 1 & 0 & 0 & 0 & 0 & 0 & 0  \\
   0 & 0 & 1 & \Delta t & 0 & 0 & 0 & 0  \\
   0 & 0 & 0 & 1 & 0 & 0 & 0 & 0  \\
   0 & 0 & 0 & 0 & 1 & \Delta t & 0 & 0  \\
   0 & 0 & 0 & 0 & 0 & 1 & 0 & 0  \\
   0 & 0 & 0 & 0 & 0 & 0 & 1 & \Delta t  \\
   0 & 0 & 0 & 0 & 0 & 0 & 0 & 1  \\
\end{matrix} \right]
\label{eqn:fk_matrix}
\end{equation}

where each non-zero element above the diagonal in each column of the matrix defines the time $\Delta t$ between the states. The measurement matrix, $H$ is then initiated as:
\begin{equation}
{{H}_{k}}=\left[ \begin{matrix}
   1 & 0 & 0 & 0 & 0 & 0 & 0 & 0  \\
   0 & 0 & 1 & 0 & 0 & 0 & 0 & 0  \\
   0 & 0 & 0 & 0 & 1 & 0 & 0 & 0  \\
   0 & 0 & 0 & 0 & 0 & 0 & 1 & 0  \\
\end{matrix} \right]
\label{eqn:hk_matrix}
\end{equation}

where a non-zero value represents the elements of which we want to measure and estimate. Thereafter, we define the state uncertainty matrix, $R_{k}={{k}_{1}}{{I}_{nxn}}$, where ${{k}_{1}}$ is the uncertainty factor and $n$ is the number of parameters that we want to estimate, in this case, is four. Lastly, we define the process noise matrix, $Q_{k}={{k}_{2}}{{I}_{2nx2n}}$, where ${{k}_{2}}$ is a constant determining the magnitude of the process noise.

With this definition, we are assuming a constant velocity and ignoring acceleration. As the controller signal to the drone is based on the estimated data provided by the Kalman filter, if the detected marker moves rapidly in the image frame and detection suddenly loses for several frames, it will follow a linear motion proportional to the estimated velocity. When the desired marker was not detected for a number of frames in a row, we reduced the speed exponentially for each state element: 
\begin{equation}
{{{\hat{x}}'}_{{{k}_{n}}}}\left( {{v}_{k}} \right)={{{\hat{x}}'}_{{{k}_{n}}}}\left( {{v}_{k-1}} \right)\alpha 
\end{equation}
where ${{k}_{n}}\le {{n}_{\max }}=8$ is denote a time step with an undetected marker, $\alpha =0.85$ denote the diminishing factor.

% Add figure
\begin{figure*}[ht!]
\centering
\includegraphics[width=\textwidth]{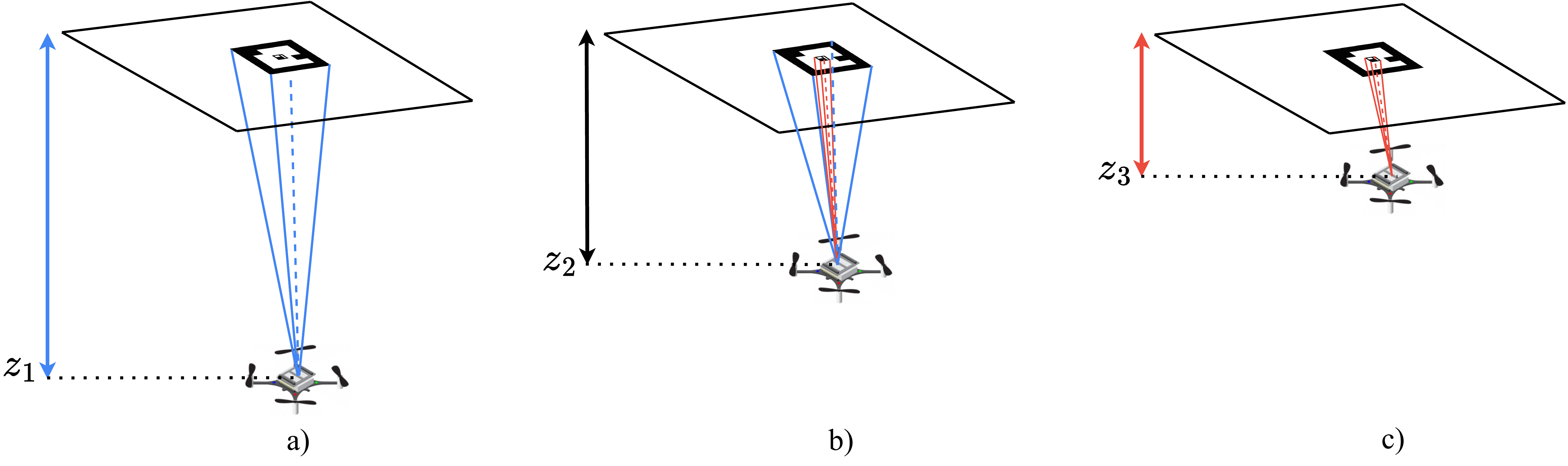}
\vspace*{-7mm}
\caption{Three stages of perching target pose estimation.}
\vspace{-5mm}
\label{fig:3_stage_markers}
\end{figure*}

\subsubsection{\textbf{Pose data merging}}

Three stages of pose estimation are indicated in Fig. \ref{fig:3_stage_markers}. The relative pose of large marker ($M_{1}$) and the small marker ($M_{2}$) to the drone are defined as ${}^{c}P_{M_{1}}={{\left[ {}^{c}{{x}_{{{M}_{1}}}},{}^{c}{{y}_{{{M}_{1}}}},{}^{c}{{z}_{{{M}_{1}}}},{}^{c}{{\psi}_{{{M}_{1}}}} \right]}^{T}}$ and ${}^{c}{{P}_{{{M}_{2}}}}={{\left[ {}^{c}{{x}_{{{M}_{2}}}},{}^{c}{{y}_{{{M}_{2}}}},{}^{c}{{z}_{{{M}_{2}}}},{}^{c}{{\psi}_{{{M}_{2}}}} \right]}^{T}}$, respectively. The process of estimating the drone's relative pose to the target, $e_{CF}={{\left[ e_{x},e_{y},e_{z},e_{\psi} \right]}^{T}}$, is expressed as below.

\textbf{\textit{Stage 1:}} Only large marker ($M_{1}$) is detected (Fig. \ref{fig:3_stage_markers}a):
\begin{equation}
{{S}_{1}}=\left\{ ({{z}_{1}},{{z}_{2}},{{z}_{CF}})\in {{\mathbb{R}}^{3}}:{{z}_{2}}\le {{z}_{CF}}\le {{z}_{1}} \right\}
\label{eqn:s1}
\end{equation}

The pose of the drone to the perching target is calculated as
\begin{equation}
e_{CF}={}^{c}{{P}_{{{M}_{1}}}}
\label{eqn:s1_pose}
\end{equation}

\textbf{\textit{Stage 2:}} Both large marker ($M_{1}$) and small marker ($M_{2}$) are detected (Fig. \ref{fig:3_stage_markers}b):
\begin{equation}
{{S}_{2}}=\left\{ ({{z}_{2}},{{z}_{3}},{{z}_{CF}})\in {{\mathbb{R}}^{3}}:{{z}_{3}}\le {{z}_{CF}}\le {{z}_{2}} \right\}
\label{eqn:s2}
\end{equation}

Let define the pose vector of ($M_{1})$ and ($M_{2}$) as below:
\begin{equation}
{}^{c}{{P}_{{{M}_{12}}}}={{\left[ {}^{c}{{x}_{{{M}_{1}}}},{}^{c}{{x}_{{{M}_{2}}}},{}^{c}{{y}_{{{M}_{1}}}},{}^{c}{{y}_{{{M}_{2}}}},{}^{c}{{z}_{{{M}_{1}}}},{}^{c}{{z}_{{{M}_{2}}}},{}^{c}{{\psi }_{{{M}_{1}}}},{}^{c}{{\psi }_{{{M}_{2}}}} \right]}^{T}}
\label{eqn:pose_m12}
\end{equation}

We obtain the pose of the drone to the perching target as
\begin{equation}
e_{CF}=W{}^{c}{{P}_{{{M}_{12}}}}
\label{eqn:s2_pose}
\end{equation}

where
\begin{equation}
W=\left[ \begin{matrix}
   {{\omega }_{x}} & 1-{{\omega }_{x}} & 0 & 0 & 0 & 0 & 0 & 0  \\
   0 & 0 & {{\omega }_{y}} & 1-{{\omega }_{y}} & 0 & 0 & 0 & 0  \\
   0 & 0 & 0 & 0 & {{\omega }_{z}} & 1-{{\omega }_{z}} & 0 & 0  \\
   0 & 0 & 0 & 0 & 0 & 0 & {{\omega }_{\psi }} & 1-{{\omega }_{\psi }}  \\
\end{matrix} \right]
\label{eqn:W}
\end{equation}

${{\omega }_{x}},{{\omega }_{y}},{{\omega }_{z}},{{\omega }_{\Psi }}$ is chosen by applying Least Mean Square (LMS) \cite{haykin2003least} filtering algorithm that minimizes cost function between the actual pose and the estimated pose of large marker ($M_{1}$) and small marker ($M_{2}$) from the camera:
\begin{equation}
C(n)=E\{|e(n){{|}^{2}}\}
\label{eqn:lms}
\end{equation}

where $e(n)$ is the error at the current state $n$ and $E\{\cdot\}$ denotes the expected value.

\textbf{\textit{Stage 3:}} Only small marker ($M_{2}$) is detected (Fig. \ref{fig:3_stage_markers}c):
\begin{equation}
{{S}_{3}}=\left\{ ({{z}_{3}},{{z}_{CF}})\in {{\mathbb{R}}^{3}}:{{z}_{CF}}\le {{z}_{3}} \right\}
\label{eqn:s3}
\end{equation}

The pose of the drone to the perching target is determined as:
\begin{equation}
e_{CF}={}^{c}{{P}_{{{M}_{2}}}}
\label{eqn:s3_pose}
\end{equation}

\subsection{Autonomous perching planner}
In  order  to  achieve  the  perching  task  in  a  safe  and  precise manner,  it is necessary to have a planner. The planner ensures the vehicle follows a perching procedure illustrated in Fig. \ref{fig:planner} which consists of seven phases:

\textit{\textbf{Phase 1:} Perching area approach}
\begin{itemize}
\item In this phase, the drone flies to reach the perching area and hovers to maintain the position as well as altitude for searching the perching target (PT).
\end{itemize}

\textit{\textbf{Phase 2:} Perching target searching}
\begin{itemize}
\item The PT is searched by markers detection until the maximum searching attempts are reached.
\item If cannot detect the marker, the vehicle will increase the altitude to gain the ability to look for it. 
\item When exceeding the trying time, the drone will be landing at the current position for safety.
\end{itemize}

\textit{\textbf{Phase 3:} Perching target pose estimation}
\begin{itemize}
\item Once the markers are detected, the proposed perching target pose estimation is utilized to get the relative pose of the drone to the PT.
\end{itemize}

\textit{\textbf{Phase 4:} Drone’s pose alignment}
\begin{itemize}
\item After having the pose, the planner will align the drone in terms of yaw-angle and horizontal until they satisfy a given condition.
\end{itemize}

\textit{\textbf{Phase 5:} Ascend to perching target}
\begin{itemize}
\item When  the  drone  is  horizontally  close  enough to  the target and the heading angle is zero, it starts ascending while maintaining its horizontal position.
\item During \textit{Phase 4} and \textit{Phase 5}, the PT pose estimation must repeat for updating the pose of the drone to the PT.
\end{itemize}

\textit{\textbf{Phase 6:} Perching execution}
\begin{itemize}
\item When the quadcopter is vertically close to the PT, the planner will operate the perching execution, including increasing the vehicle  throttle  (so  that  the magnetic perching gear can mount quickly  to  the perching center) and disabling stabilized control.
\end{itemize}

\textit{\textbf{Phase 7:} Perching completion}
\begin{itemize}
\item Once the vehicle is perched, that means it is stationary hanging at the magnetic perching point, the planner will turn off its motors, and the perching task is completed.
\item If the perching task fails and the predefined maximum perching attempts do not exceed, the drone will descend 10cm and go back to \textit{Phase 3} to try again.
\item Once the perching time has passed, the drone will land at its current position to prevent the lack of battery.
\end{itemize}

% Add figure
\begin{figure}[ht!]
\centering
\includegraphics[width=8cm]{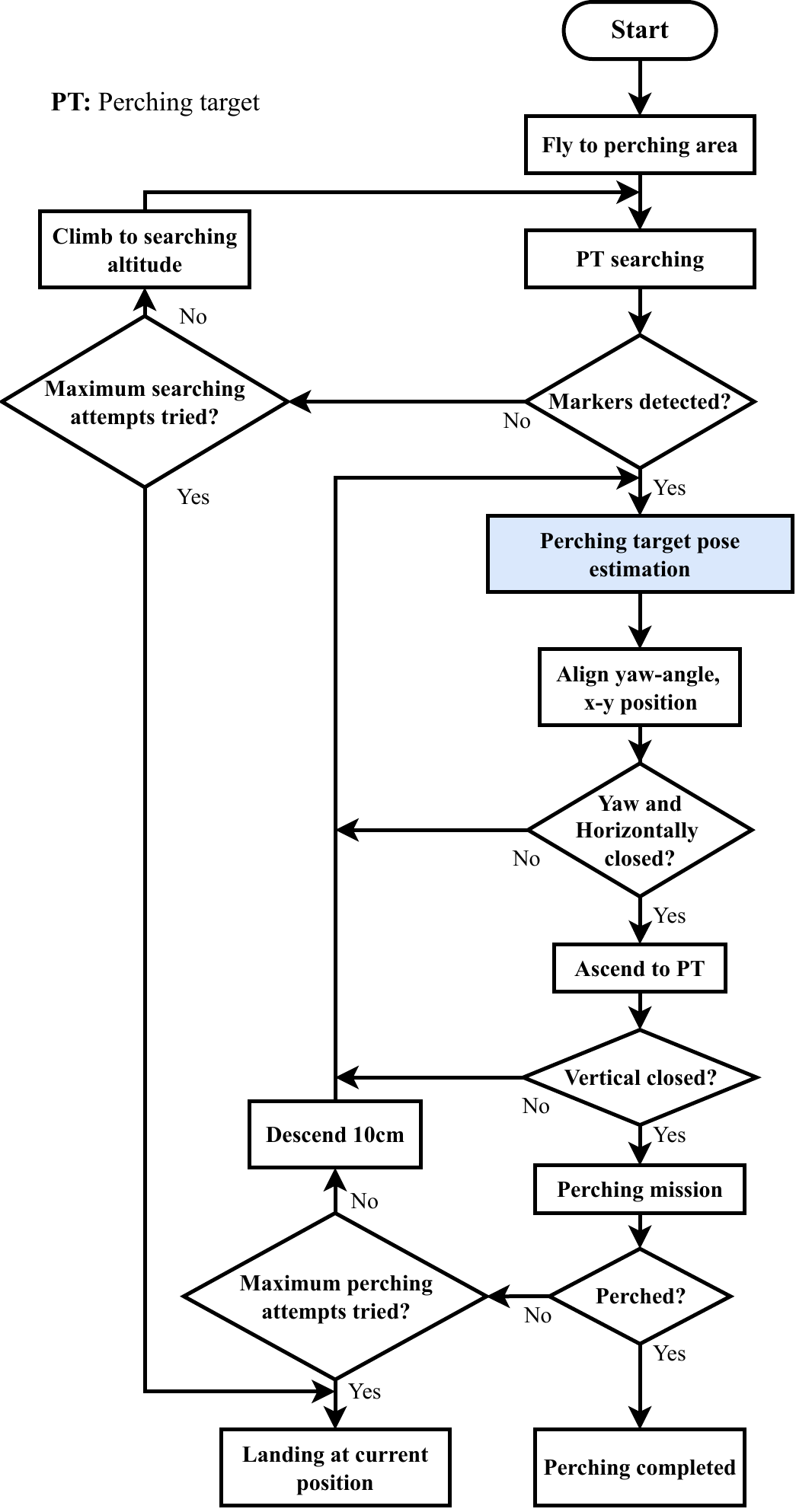}
\vspace*{-2mm}
\caption{The flowchart of the autonomous precision perching procedure.}
\label{fig:planner}
\end{figure}

\section{Experiments} \label{Experiments}
\subsection{Systems Configurations}
The experiments were conducted with the VICON Motion Capture Positioning System consisting of eight cameras as illustrated in Fig. \ref{fig:testbed}a). On the perching plane, the designed marker is printed and placed downward. The perching target of multi-markers is equipped with a round magnet with a diameter of $50mm$.

The drone used in this research is a Crazyflie 2.1 nano UAV. The FPV camera is mounted on a drone and pointed upwards in an eye-in-hand configuration for detecting visual markers and estimating relative pose as shown in Fig. \ref{fig:testbed}b). Image frames collected from the camera have a size of 640 x 480 pixels. A small and strong magnetic perching gear is held on the top of the drone (Fig. \ref{fig:testbed}c). The total weight of constructed Crazyflie is only 42 grams, including the battery and reflective markers.

% Add figure
\begin{figure}[b!]
\centering
\includegraphics[width=7.5cm]{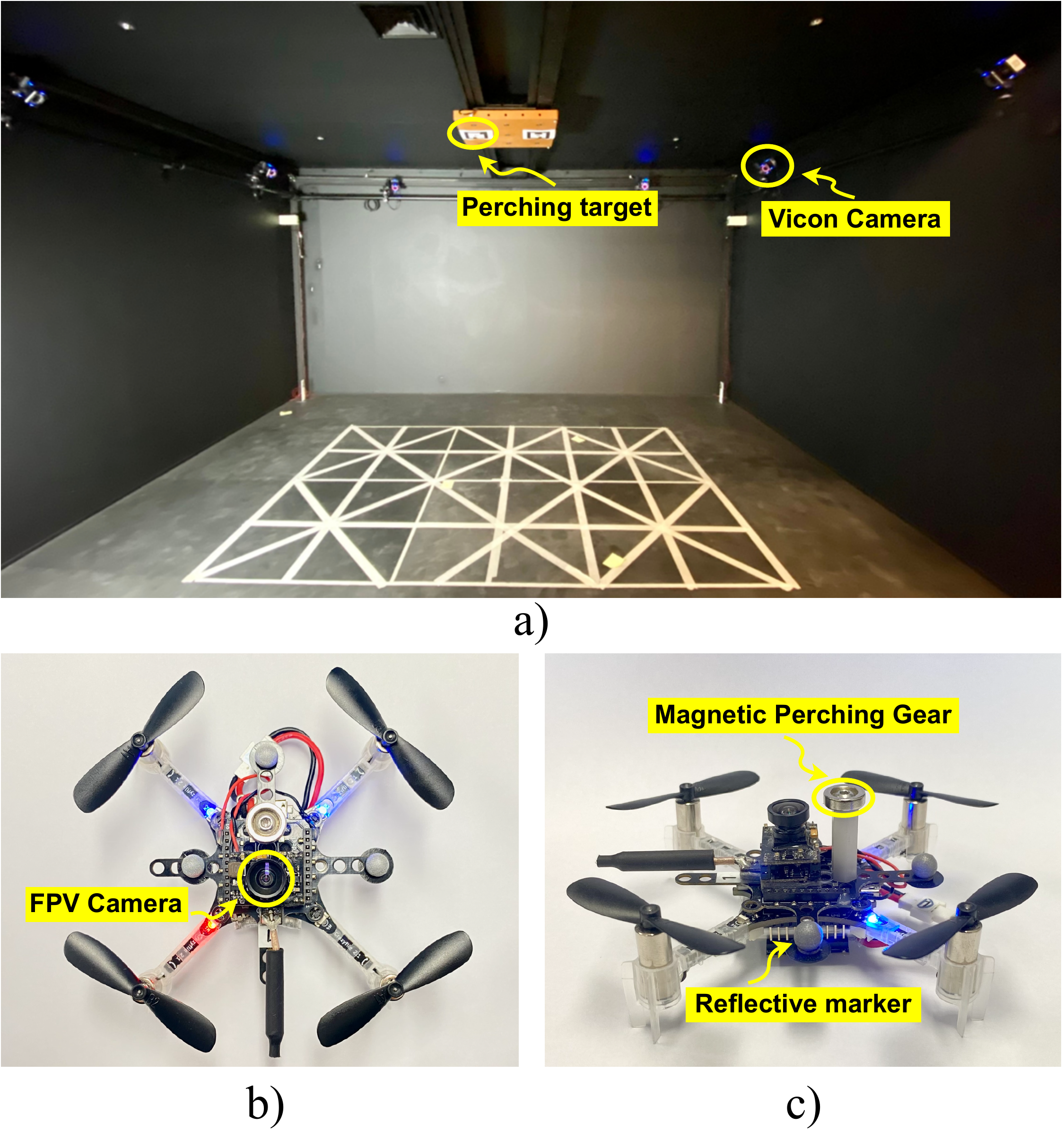}
\vspace*{-2mm}
\caption{Experimental conditions.}
\centering
\label{fig:testbed}
\end{figure}
% Add figure
\begin{figure}[b!]
\centering
\includegraphics[width=8cm]{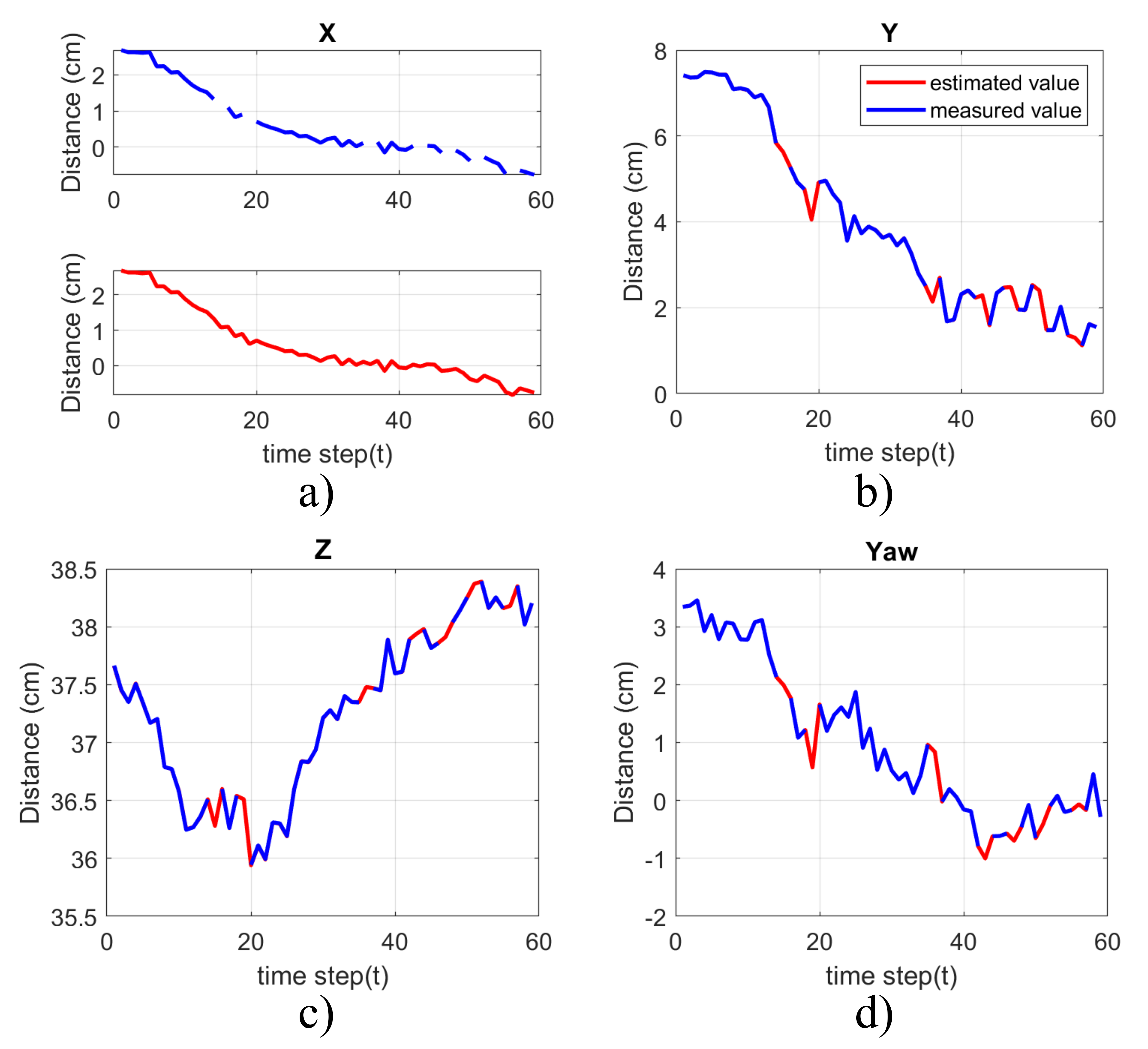}
\vspace*{-2mm}
\caption{The target pose data after utilizing the Kalman filter.}
\centering
\label{fig:kalman_results}
\end{figure}

\subsection{Experimental results}

Our experiments indicate that the large $(M_{1})$ and small $(M_{2})$ markers begin to be detected at a maximum distance $z_{1}=115$cm, $z_{2}=25$cm, based on the experimental measurements performed. In the absence of a large marker, the minimum relative distance is $z_{3}=12$cm.

Figure \ref{fig:kalman_results} shows plots over a short tracking sequence where the drone was moved in several different positions. In the real cases of perching target pose calculation, the output data is not consecutive without the Kalman filter estimation. This will lead to drone crashes. The red line represents the estimated values obtained after using the Kalman filter, whereas the blue line shows the raw data output from the ArUco detection and pose estimation function of OpenCV. As can be seen upon inspection of the figure, during the measurement period, the Kalman filter is capable of making a good estimate of the movement when the measurement input data is lost. Especially at $t\approx18$s, no marker is detected for approximately one second, and Kalman filters predict marker movement using velocity at this time, decreasing the velocity by the factor $\alpha = 0.85$ for each iteration until the maximum number of frames $n_{max} = 8$ is reached.

To determine the values of the merging coefficient, the estimated values and actual values of them were collected at ten random positions and directions. There are $100$ sequential data values at each point. The LMS algorithm is then used to find the optimal coefficient values with the least amount of error between actual and estimated data. We achieve ${{\omega}_{x}}=0.275$, ${{\omega}_{y}}=0.306$, ${{\omega}_{z}}=0.728$, ${{\omega}_{\psi}}=0.469$.

Table \ref{tb1} shows relative position data $(e_{x},e_{y},e_{z})$ including the actual values, estimated values, and the peak-to-peak magnitudes of the error between them in centimeters. We measured and estimated the position values at five different points of the drone, the peak-to-peak estimated position error magnitudes in the range of 0.6cm and decreasing when the drone approaches near and close to the center of the target.

In addition, we also gathered the relative heading angle in degree $(e_{\psi})$. There is no significant difference between actual values and estimated values as shown in Table \ref{tb2}. When the $yaw$-angle is close to the straightforward direction, the rounded error becomes zero.

\renewcommand{\arraystretch}{1.5} % To change the height of table row
\begin{table}[!t]
\centering
\caption{The estimation of relative position values in centimeters.}\label{tb1}
\resizebox{\columnwidth}{!}{%
\begin{tabular}{|c|c|c|c|c|c|}
\hline
\textbf{\begin{tabular}[c]{@{}c@{}}Actual \\ Position\end{tabular}}    & (0, 0, 24)        & (4, 6, 16)       & (5, -7, 11)        & (-10, 11, 7)        & (-6, -3, 18)        \\ \hline
\textbf{\begin{tabular}[c]{@{}c@{}}Estimated \\ Position\end{tabular}} & (0.4, -0.4, 24.5) & (4.4, 5.7, 16.4) & (5.3, -7.4, 11.3)  & (-10.5, 11.6, 7.2)  & (-6.3, -3.3, 18.5)  \\ \hline
\textbf{\begin{tabular}[c]{@{}c@{}}Est. Position\\ Error\end{tabular}} & (0.4, -0.4, 0.5)  & (0.4, -0.3, 0.4)  & (0.3, -0.4, 0.3)    & (-0.5, 0.6, 0.2)     & (-0.3, -0.3, 0.5) \\ \hline
\end{tabular}%
}
\end{table}

\begin{table}[!t]
\centering
\caption{The estimation of relative heading angle values in degree.}\label{tb2}
\resizebox{0.8\columnwidth}{!}{%
\begin{tabular}{|
% % >{}c |c|c|c|c|c|c|c|c|}
>{}c |m{0.7cm}|m{0.7cm}|m{0.7cm}|m{0.7cm}|m{0.7cm}|m{0.7cm}|m{0.7cm}|m{0.7cm}|}
\hline
\textbf{\begin{tabular}[c]{@{}c@{}}Actual \\ Heading Angle\end{tabular}}    & \hfil175 & \hfil-165 & \hfil125 & \hfil-90 & \hfil30 & \hfil-15 & \hfil5 & \hfil1 \\ \hline
\textbf{\begin{tabular}[c]{@{}c@{}}Estimated \\ Heading Angle\end{tabular}} & \hfil172 & \hfil-167 & \hfil128 & \hfil-92 & \hfil32 & \hfil-14  & \hfil5 & \hfil1 \\ \hline
\textbf{\begin{tabular}[c]{@{}c@{}}Est. Heading Angle\\ Error\end{tabular}} & \hfil-3   & \hfil-2   & \hfil3   & \hfil-2  & \hfil2  & \hfil1  & \hfil0 & \hfil0 \\ \hline
\end{tabular}%
}
\end{table}

The real flight test video can be accessed here: \url{http://bit.ly/auto\_perching\_gnc\_2304}.
% \textit{http://bit.ly/auto\_perching\_gnc\_2304}.

% https://ieeexplore.ieee.org/abstract/document/9852196

\section{Conclusion} \label{Conclusion}

This research proposed a vision-based autonomous perching approach for nano quadcopters onto a predefined perching target on horizontal surfaces. A monocular camera mounted on Crazyflie captured the image frame of the multi-marker used as a perching target. After that, the markers will be detected and the pose calculated. Kalman filters were applied to fill in the empty gaps caused by missing transmissions or detection failures. Then the pose from multiple markers was combined with the optimal coefficients achieved from the Least Mean Squared algorithm. Finally, the autonomous perching planner was used to perform the real flight test with the VICON positioning system. Experimental results have confirmed the efficiency and practicality of the presented approach. 

There are still directions for further development to achieve even more remarkable results. In future works, autonomous swarm perching of nano drones will be conducted.

\section*{Acknowledgment}

This research was supported by the MSIT (Ministry of Science and ICT), Korea, under the ITRC (Information Technology Research Center) support program (IITP-2023-2018-0-01423) supervised by the IITP (Institute for Information \& Communications Technology Planning \& Evaluation).

This research was supported by Basic Science Research Program through the National Research Foundation of Korea (NRF) funded by the Ministry of Education (2020R1A6A1A03038540). 

This work was supported by Future Space Navigation \& Satellite Research Center through the National Research Foundation funded by the Ministry of Science and ICT, the Republic of Korea (2022M1A3C2074404).
%%%%%%%%%%%%%%%%%%%%%%%%%%%%%%%%%%%%%%%%%%%%%%%%%%%%%%%%%%%%%%%%%%%%%%%%%%%%%%%%
\balance
\bibliographystyle{ieeetr}
\bibliography{references}

\end{document}